\documentclass{article}
\usepackage{spconf,amsmath,epsfig}

\usepackage{color}
\usepackage{tikz}
\usepackage{lipsum}
\usepackage{multirow}
\usepackage[font=scriptsize]{subfig}
\usepackage[inline]{enumitem}
\usepackage{xfrac}
\usepackage{booktabs}
\usepackage{hyperref}

\usepackage{stackengine}
\usepackage{multicol}
\usepackage{array}
\newcolumntype{P}[1]{>{\centering\arraybackslash}p{#1}}

\newcommand{\modif}[1]{#1}
\newcommand{\newtext}[1]{#1}
\newcommand{\newtextt}[1]{#1}

\usepackage[absolute,showboxes]{textpos}

\TPMargin{5pt}

\newcommand{\copyrightstatement}{
	\begin{textblock}{0.84}(0.08,0.93)    
		\noindent
		\scriptsize
		\copyright 2019 IEEE. Published in the IEEE 2019 International Geoscience \& Remote Sensing Symposium (IGARSS 2019), scheduled for July 28 - August 2, 2019 in Yokohama, Japan. Personal use of this material is permitted. However, permission to reprint/republish this material for advertising or promotional purposes or for creating new collective works for resale or redistribution to servers or lists, or to reuse any copyrighted component of this work in other works, must be obtained from the IEEE. Contact: Manager, Copyrights and Permissions / IEEE Service Center / 445 Hoes Lane / P.O. Box 1331 / Piscataway, NJ 08855-1331, USA. Telephone: + Intl. 908-562-3966.
	\end{textblock}
}

\title{Robust Building-based Registration of Airborne LiDAR Data and Optical Imagery on Urban Scenes}
\name{Thanh Huy Nguyen$^{\ast \dagger}$, Sylvie Daniel$^{\ast}$, Didier Gu\'{e}riot$^{\dagger}$, Christophe Sint\`{e}s$^{\dagger}$ and Jean-Marc Le Caillec$^{\dagger}$ \thanks{Thanks to the Natural Sciences and Engineering Research Council of Canada and the Brittany region (France) for funding this project.}
}

\address{$^{\ast}$Universit\'{e} Laval, Qu\'{e}bec City, QC G1V 0A6, Canada\\
	$^{\dagger}$IMT Atlantique, UMR 6285 CNRS LabSTICC, 29238 Brest Cedex 3, France\\[2.5pt]
	{\small \texttt{\{thanh.nguyen, didier.gueriot, christophe.sintes, jm.lecaillec\}@imt-atlantique.fr,}}\\
	{\small \texttt{sylvie.daniel@scg.ulaval.ca}}}

\begin{document}
	\copyrightstatement
	
	\ninept
	\maketitle
	\begin{abstract}
		The motivation of this paper is to address the problem of registering airborne LiDAR data and optical aerial or satellite imagery acquired from different platforms, at different times, with different points of view and levels of detail. In this paper, we present a robust registration method based on building regions, which are extracted from optical images using mean shift segmentation, and from LiDAR data using a 3D point cloud filtering process. The matching of the extracted building segments is then carried out using Graph Transformation Matching (GTM) which allows to determine a common pattern of relative positions of segment centers. Thanks to this registration, the relative shifts between the data sets are significantly reduced, which enables a subsequent fine registration and a resulting high-quality data fusion.
	\end{abstract}
	%

	\section{Introduction}
	\label{sec:intro}
	Over the years, existing works in the domain of aerial or satellite imagery and airborne LiDAR fusion have addressed very specific acquisition contexts, \modif{where} the respective image and LiDAR 3D point cloud are already registered and/or they are acquired from the same platform at identical or very close dates. For example, solutions proposed in the {2013 GRSS Data Fusion Contest} \cite{debes2014hyperspectral} focused on performing a fusion between LiDAR data and hyperspectral imagery with the same spatial resolution and acquisition dates on two consecutive days. In 2015, the contest \cite{vo2016processing} involved extremely high resolution LiDAR data and RGB imagery collected from the same airplane with the sensors rigidly fixed to the same platform. Thus, the solutions submitted to these contests have never been intended to overcome the inherent obstacles of data sets collected from different platforms with different acquisition configuration (e.g. flying track, height, orientation, etc.) at different moments and even in different seasons, with different spatial resolutions and levels of detail. The need for a relevant registration in such context is exemplified in the work undertaken by Cura et al. \cite{cura2017scalable}. \modif{It} relates to the rise of Geographical Information System (GIS) availability, in particular through the open data movement, that requires the integration of data from multiple and heterogeneous sources. However, a solution that is versatile enough to satisfy this difficult context still remains an unsolved research problem.
	
	Accurate registration of LiDAR data and optical imagery is a prerequisite to data fusion applications \cite{Parmehr2014}. The majority of automatic methods for registration of such data sets can be classified into two categories, namely intensity-based and feature-based methods. 
	Feature-based methods establish correspondence between data sets based on available distinguishable features. They involve feature extraction algorithms and  feature matching strategy \cite{palenichka2010automatic}. \newtext{On the other hand}, intensity-based methods determine the optimal sensors pose by maximizing a statistical similarity (e.g. mutual information) between the values of, respectively, image pixels and LiDAR-derived pixels \cite{Parmehr2014,Mastin2009}. \newtext{However,} in addition to the \newtext{high} computational cost, the main drawback of these methods is the needs for the data sets to be spatially close to each other, to have the same resolution and to display similar intensity characteristics \cite{Parmehr2014,Mastin2009}. As a result, we present in this paper a novel feature-based registration approach capable of overcoming the challenges of the  aforementioned research context. We focus on urban scenes and more specifically on buildings as primitives on which the matching between the data sets relies. 
	
	\newtext{The paper is organized as follows. Section \ref{sec:proposed} is devoted to the description of the proposed registration method, consisting of three successive steps, namely feature extraction, feature matching and transformation model estimation. Then, experimental results involving different data sets are presented in Section \ref{sec:results}. Finally, Section \ref{sec:conclusions} provides conclusions and perspectives of this work.}
	
	
	\section{Proposed registration method}
	\label{sec:proposed}
	
\begin{figure}[!t]
		\centering
		\begin{tikzpicture}[every text node part/.style={align=center},every node/.style={scale=0.9}]
		\node[draw,fill=white,text=black,minimum height=0.35cm,text width=2.5cm,rounded corners] (Input1) at (0,-0.8) {\footnotesize LiDAR point cloud};
		\node[draw,fill=white,text=black,minimum height=0.35cm,text width=2cm,rounded corners] (Input2) at (4,-0.8) {\footnotesize Optical image};
		
		\node[draw,fill=cyan,text=black,minimum height=0.35cm,text width=2.5cm] (Process1) at (0,-1.65) {\footnotesize  Building point extraction};
		\node[draw,fill=cyan,text=black,minimum height=0.35cm,text width=2.5cm] (Process2) at (4,-1.65) {\footnotesize  Meanshift segmentation and refinement};
		
		
		\node[draw,fill=white,text=black,minimum height=0.35cm,text width=2.5cm,rounded corners] (Output1) at (0,-2.5) {\footnotesize Building 3D regions};
		\node[draw,fill=white,text=black,minimum height=0.35cm,text width=2.75cm,rounded corners] (Output2) at (4,-2.5) {\footnotesize Building 2D segments};
		
		\node[draw,fill=cyan,text=black,minimum height=0.75cm,text width=3.5cm] (Pairing) at (2,-3.5) {\footnotesize  Comparing segments and determining correspondences};
		
		\node[draw,fill=white,text=black,minimum height=0.35cm,text width=3cm,rounded corners] (Correspondences) at (2,-4.4) {\footnotesize Segment correspondences};
		
		\node[draw,fill=cyan,text=black,minimum height=0.35cm,text width=2.5cm] (TME) at (2,-5.2) {\footnotesize Transformation model estimation};
		\node[draw,fill=white,text=black,minimum height=0.35cm,text width=2.5cm,rounded corners] (Params) at (2,-6.2) {\footnotesize Estimated camera pose parameters};
		
		\draw[->,draw=black] (Input1) -- (Process1);
		\draw[->,draw=black] (Input2) -- (Process2);
		\draw[->,draw=black] (Process1) -- (Output1);
		\draw[->,draw=black] (Process2) -- (Output2);
		\draw[->,draw=black] (Output1) -- (Pairing);
		\draw[->,draw=black] (Output2) -- (Pairing);
		\draw[->,draw=black] (Pairing) -- (Correspondences);
		\draw[->,draw=black] (Correspondences) -- (TME);
		\draw[->,draw=black] (TME) -- (Params);
		\end{tikzpicture}
		\caption{Flowchart of the building-based registration between optical image and LiDAR point cloud.}
		\label{fig:flowchart_seg}
	\end{figure}
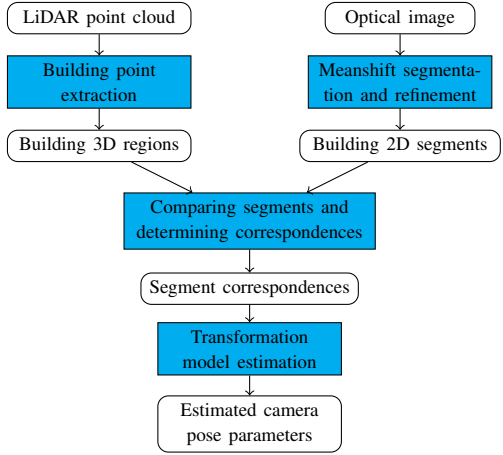

	\newtextt{Our novelty resides in a methodology that carries out effectively building extraction and matching, by the virtue of well-tailored series of well-known processes and algorithms}.
	The main steps of the proposed registration method based on buildings is illustrated by Fig. \ref{fig:flowchart_seg}. On each data set we perform different processes with the purpose of extracting buildings from the observed urban scene. On the one hand, we apply an elevation thresholding on LiDAR point cloud 3D coordinates in order to \modif{select} building points. On the other hand, mean shift segmentation is performed on the optical image with a carefully chosen bandwidth parameter, followed by a  refinement to remove unwanted segments and preserve building-like ones. 
	  
	
	\subsection{Feature extraction}
	\subsubsection{Building extraction from LiDAR data}
	\label{subsec:building_seg_lidar}
	The extraction of buildings from LiDAR point cloud is carried out through a series of steps, as follows: 
	\begin{enumerate}[label={\footnotesize \textit{Step \arabic*} $|$}]
		\itemsep0em
		\item[\footnotesize \textit{Input} $|$] LiDAR 3D point cloud $(X,Y,Z)$.
		\item Elevation thresholding: separating non-ground points from ground points depending on their elevation value. This task is proposed by many existing works as an initial but necessary step \cite{awrangjeb2013automatic}. The elevation threshold value is calculated as follows: $ T_e=\mathrm{mean}(z_G) + \max\{2.5, \mathrm{std}(z_G)\} $, where $z_G$ denotes the altitude of ground points.
		
		\item Vertical projection: all non-ground points are vertically projected onto the plan $ z=0 $, which creates a 2D binary mask of non-ground points. The resolution of this binary mask is set accordingly to the point density of the input LiDAR point cloud to avoid null-value pixels, e.g. a resolution of 1 meter $ \times $ 1 meter for a point cloud of 2 points/m$ ^2 $ density.
		\item Morphological opening is then applied on the binary mask to remove small regions as well as rounding up bigger ones. The morphological structuring element is a diamond shape.  \newtext{Its size is 5 or 7 pixels (depending on the area)}.
		\item Connectivity labeling: connecting pixels into segments based on their connectivity, and then labeling these segments.
		\item A removal of small regions that are smaller than 20 square meters is carried out, which results in a labeled building mask.
		\item Extracting building points: based on the labeled building mask, we select among the non-ground points only the regions that are seeded by labeled segments.
		\item[\footnotesize \textit{Output} $|$] Building 3D regions and their boundary.
	\end{enumerate}
	
	\subsubsection{Building segmentation from optical image using Mean shift}
	\label{subsec:building_seg_img}
	Mean shift is an unsupervised clustering method widely used in many areas of Computer Vision, including 2D shape extraction, and texture segmentation \cite{comaniciu2002mean}. Compared to $ k $-means clustering, mean shift does not require a prior number of classes, but a value of bandwidth corresponding to the image  color range and size of objects to be segmented. Moreover, in an urban area, $ k $-means fails to segment buildings because building roof color varies a lot, and also building roofs and streets may have similar color. 
	
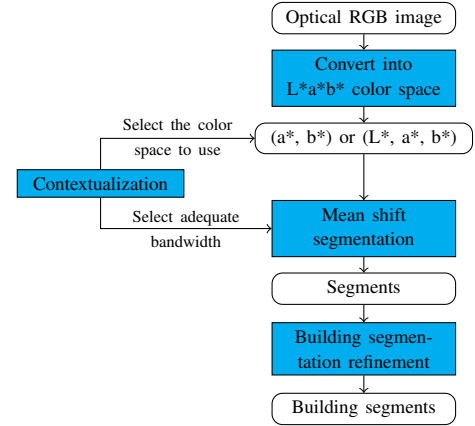
\begin{figure}[!t]
		\centering
		\begin{tikzpicture}[every text node part/.style={align=center},every node/.style={scale=0.9}]
		\node[draw,fill=white,text=black,minimum height=0.35cm,text width=2.5cm,rounded corners] (a) at (3,-0.2) {\footnotesize Optical RGB image};
		\node[draw,fill=cyan,text=black,minimum height=0.35cm,text width=2.5cm] (1) at (3,-1) {\footnotesize Convert into L*a*b* color space};
		\node[draw,fill=white,text=black,minimum height=0.35cm,text width=3cm,rounded corners] (b) at (3,-1.8) {\footnotesize (a*, b*) or (L*, a*, b*)};
		
		\node[draw,fill=cyan,text=black,minimum height=0.35cm,text width=2.25cm] (A) at (-0.5,-2.4) {\footnotesize Contextualization};
		\node[draw,fill=cyan,text=black,minimum height=0.35cm,text width=2.5cm] (2) at (3,-3) {\footnotesize Mean shift segmentation};
		\node[draw,fill=white,text=black,minimum height=0.35cm,text width=2.5cm,rounded corners] (c) at (3,-3.8) {\footnotesize Segments};
		\node[draw,fill=cyan,text=black,minimum height=0.35cm,text width=2.5cm] (3) at (3,-4.6) {\footnotesize Building segmentation refinement};
		\node[draw,fill=white,text=black,minimum height=0.35cm,text width=2.5cm,rounded corners] (d) at (3,-5.4) {\footnotesize Building segments};
		\draw[->,draw=black] (a) -- (1);
		\draw[->,draw=black] (1) -- (b);
		\draw[->,draw=black] (b) -- (2);
		\draw[->,draw=black] (A) |- node[near end,text width=2cm]{\scriptsize Select adequate bandwidth} (2);
		\draw[->,draw=black] (A) |- node[near end,text width=1.75cm]{\scriptsize Select the color space to use} (b);
		\draw[->,draw=black] (2) -- (c);
		\draw[->,draw=black] (c) -- (3);
		\draw[->,draw=black] (3) -- (d);
		
		\end{tikzpicture}
		\caption{Flowchart of the building segmentation from optical image.}
		\label{fig:flowchart_seg_img}
	\end{figure}

	
	Fig. \ref{fig:flowchart_seg_img} presents a flowchart of the building segmentation on optical image using mean shift algorithm.
	First, the optical visible image is converted into the CIE L*a*b* color space, as this color space allows better distinction of objects than RGB color space. 
	For the satellite imagery, a pansharpening 
	is carried out to merge 50-cm resolution panchromatic and 2-m resolution multispectral imagery to create a 50-cm color image, which will be segmented by mean shift. 
	However, determining the best bandwidth parameter for mean shift still remains difficult even though a number of approaches have been explored \cite{chacon2013comparison}. Thus, this bandwidth parameter should be set manually according to the type of area (either residential, industrial, mixed, etc.), and the size of objects of interest. In other words, the bandwidth parameter selection is based on the contextualization of the scene, alongside with the choice between (a*, b*) values and (L*, a*, b*) values.
	
\begin{figure}[t]
		\subfloat[On a tree segment]{\includegraphics[height=2.5cm]{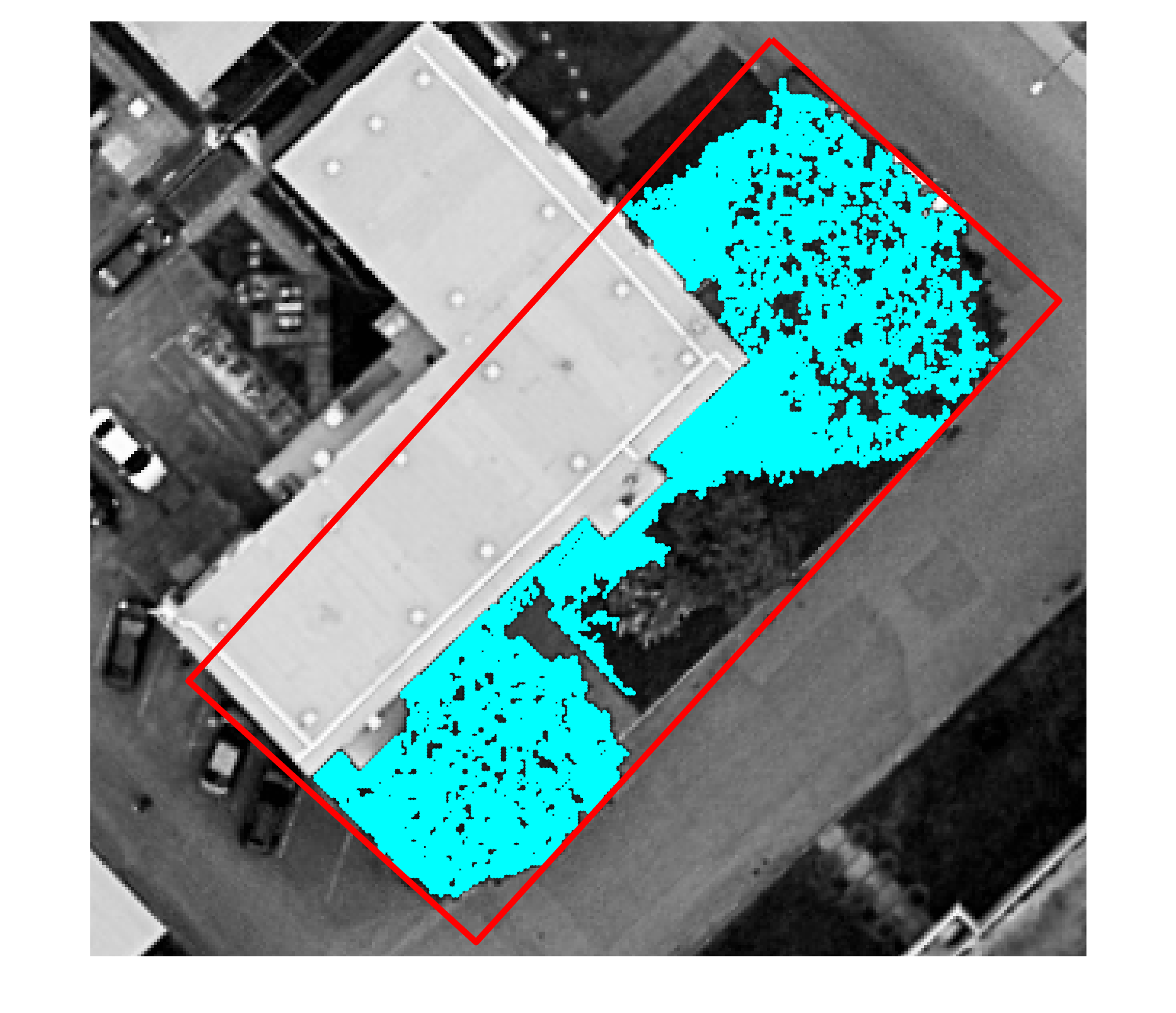}}\hfill
		\subfloat[On building segments]{\includegraphics[height=2.5cm]{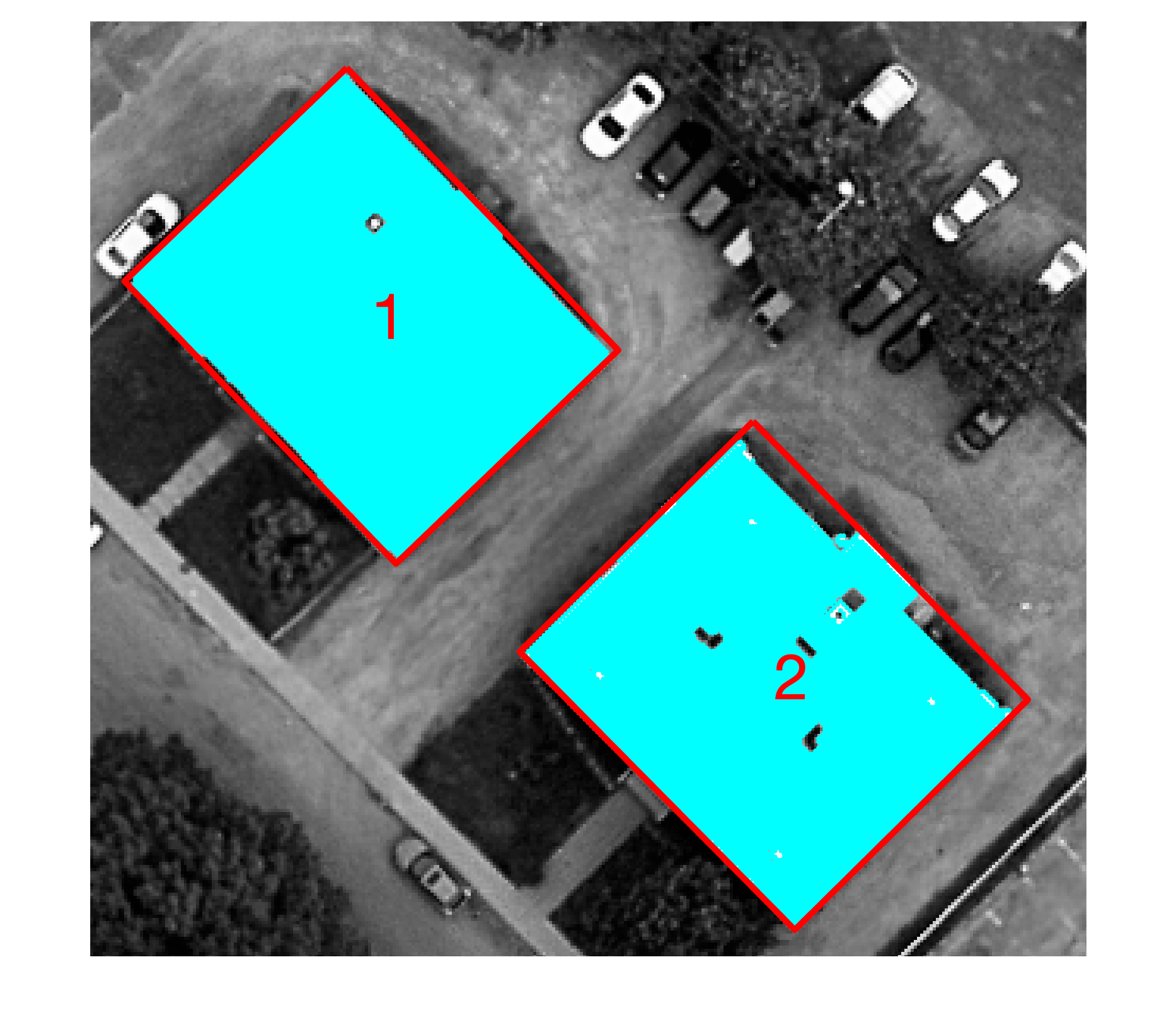}}\hfill
		\subfloat[MBR filling percentage]{
			\scalebox{0.7}{%
				\begin{tabular}[b]{c|c}
					\hline
					\textbf{Segment} & \textbf{\%\textsubscript{MBR filling}} \\
					\hline
					Tree & 43.41\%\\
					\hline
					Building 1 &   94.75\%\\
					Building 2 & 91.57\%\\
					\hline
				\end{tabular}
				\label{tab:percent}
			}
		}
		\vspace{-.25cm}
		\caption{Illustration of MBR (in red) of a tree segment versus building segments (cyan points) \newtextt{and comparison of their filling percentage.}}
		\label{fig:MBR_LER}
	\end{figure}

	When applying the mean shift segmentation, many building regions are segmented alongside with other regions related to trees, streets, or cars. 
	Obviously, these unwanted non-building segments need to be removed before \modif{carrying out} the comparison with the building segments extracted from LiDAR point cloud. To this purpose, we \newtext{rely on the number of pixels inside each segment and remove small segments since they usually correspond to trees and cars. Large segments are removed similarly since they correspond to street regions.} This filter is simple and efficient \cite{gavankar2018automatic}, but completely dependent on the image resolution. Therefore, it needs a manual intervention to be set correctly.
	The authors of \cite{gavankar2018automatic} also proposed two additional filters based on the length ratio of the segment major and minor axis, and the segment eccentricity to remove falsely detected building segments and keep the segments that are associated to rectangular and round building regions. However, they are not effective in the case of complex building segments. 
	\newtext{Also, it is not clear how the axes and the eccentricity of the segments are determined}. In addition, the thresholds used by these filters are highly subjective. In this paper, we present another approach to discriminate buildings apart from regions that relate to trees or streets. After applying the preliminary filter based on the number of pixels inside each segment, we identify the minimal bounding rectangle (MBR) of each segment.  Based on this rectangle, we calculate the filling percentage of each segment $\%_{\textnormal{MBR filling}}=\sfrac{\textnormal{Area(segment)}}{\textnormal{Area(MBR)}}\times 100$. This percentage is then used to filter the unwanted segments, as filling percentage of a rectangle building segment should be higher than that of an unwanted segment, as compared on Fig. \ref{fig:MBR_LER}.
	
	
	\begin{figure}[t]
		\centering
		\subfloat[LiDAR building segments]{\includegraphics[trim=4cm 2.5cm 3cm 1.75cm,clip,height=3.5cm]{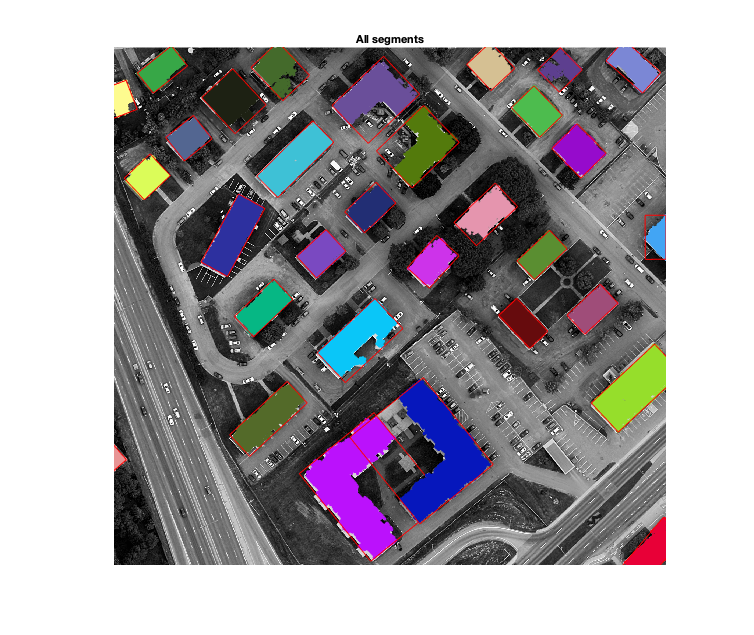}}\hspace{0.25cm}
		\subfloat[Image building segments (MBR filling percentages $ >50\% $)]{\includegraphics[trim=0.5cm 0.5cm 0.5cm 0.5cm,clip,height=3.5cm]{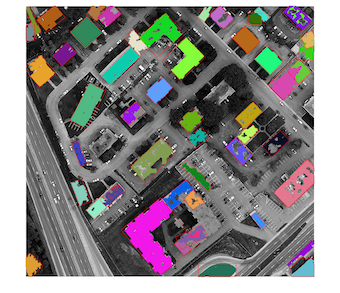}}
		\vspace{-.25cm}
		\caption{Extracted building segments from optical image segmentation and from LiDAR point cloud.}
		\label{fig:seg_to_compare}
	\end{figure}
	
	\subsection{Feature comparison and matching}
	After extracting building segments from the data sets, the next step is to compare and match them. From the optical image, we select the segments that have a MBR filling percentage higher than 50\%. 
	On the other hand, all building regions extracted from the LiDAR point cloud will be taken into consideration. Both sets of extracted segments are depicted on Fig. \ref{fig:seg_to_compare}.
	
	Segment comparison and matching issues are anticipated. Indeed, the data sets are relatively distant to each other (cf. Table \ref{tab:datasets}), and wrongly extracted segments may still exist after the MBR-based segment refinement. Therefore, matching the segments based on their spatial relation w.r.t. their neighbors is more relevant than comparing their individual values. 
	Indeed, taking into account the position of the segment center, a common pattern representing the relative spatial arrangement of the data sets can be determined using GTM (Graph Transformation Matching) algorithm \cite{aguilar2009robust}. GTM is a graph-based point matching algorithm designed for solving the registration between images with non-rigid deformations. This algorithm performs better than RANSAC in removing outliers on test image data sets \cite{aguilar2009robust}, as well as in our work \newtextt{(cf. Table \ref{tab:comparative})}. 
	\begin{figure}[t]
		\centering
		\subfloat[Initial matching]{\includegraphics[trim=3cm 1.5cm 3cm 1.5cm,clip,height=3.5cm]{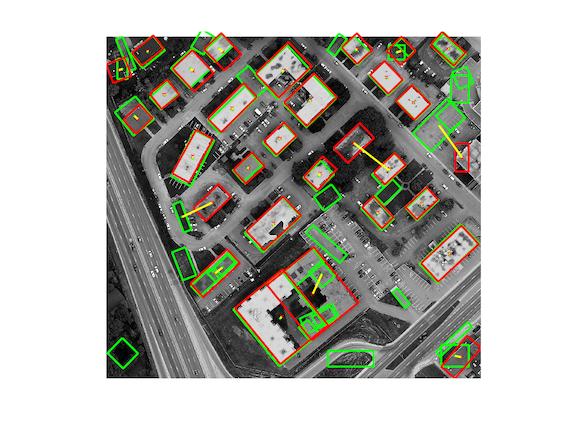}\label{subfig:initial}}\hspace{0.02cm}
		\subfloat[RANSAC result]{\includegraphics[trim=3cm 1.5cm 3cm 1.5cm,clip,height=3.5cm]{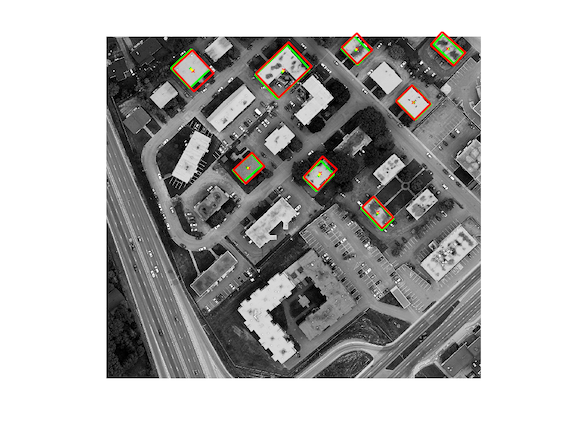}\label{subfig:ransac}}
		
		\vspace{-0.2cm}
		
		\subfloat[GTM result]{\includegraphics[trim=6cm 3cm 6cm 3cm,clip,height=3.5cm]{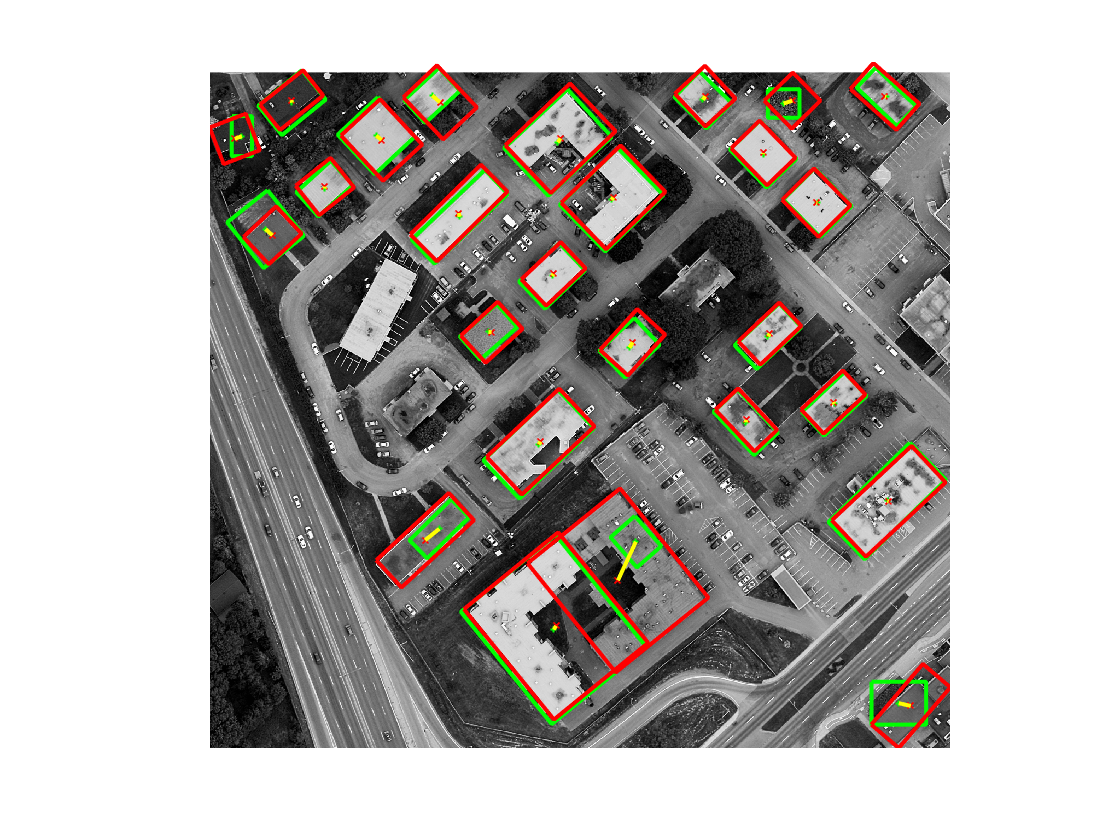}\label{subfig:gtm}}\hspace{0.1cm}
		\subfloat[GTM + Area and direction validation]{\includegraphics[trim=6cm 3cm 6cm 3cm,clip,height=3.5cm]{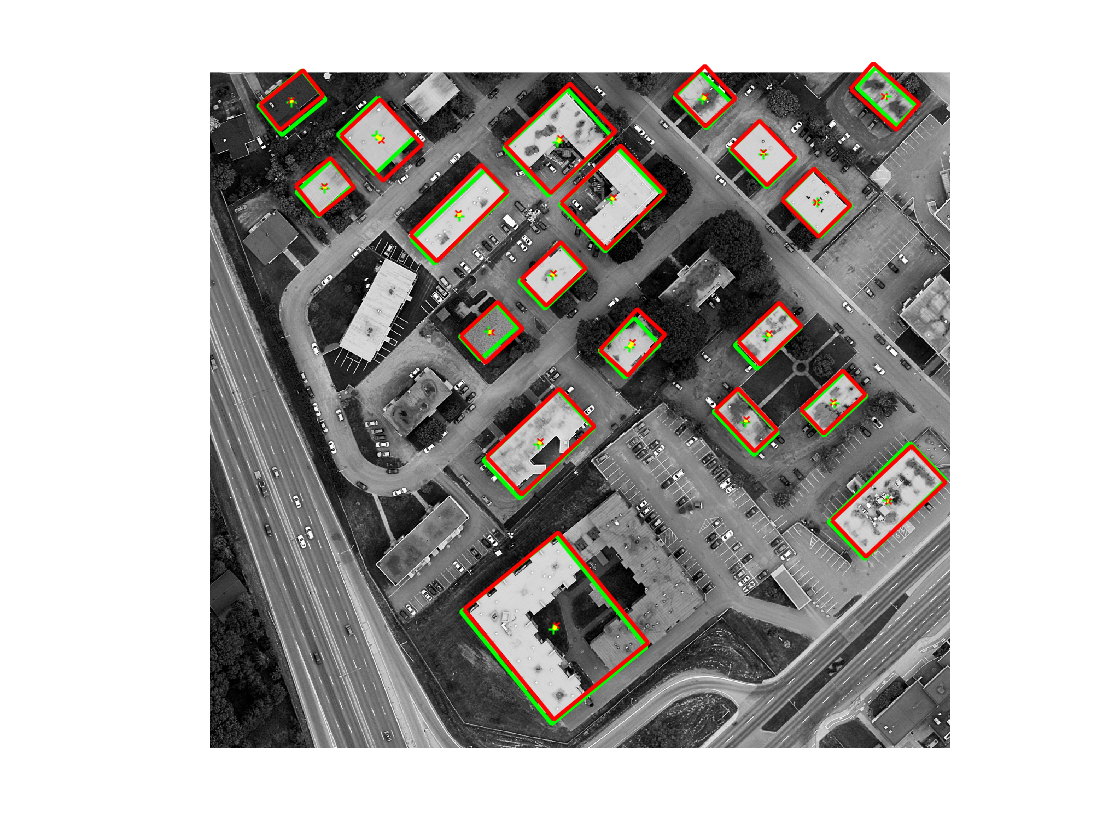}\label{subfig:gtm_final}}
		\vspace{-0.2cm}
		\caption{Considering relative position of segment centers (green and red rectangles represent the MBR of the segments extracted from optical image and LiDAR point cloud).} 
		\label{fig:GTM}
	\end{figure}
	In practice, both GTM and RANSAC require an initial one-to-one matching of segment centers, which can be carried out relying on the positions of vertically projected 3D building region centers onto plan $ z=0 $ and the centers of 2D segments (extracted by mean shift segmentation). If the relative shifts are too big (e.g. data set no. 3), this initial matching is added with a translation vector calculated based on the displacement of the largest segment. 
	The results of the initial matching of segment centers, followed by RANSAC are shown on Fig. \ref{subfig:initial} and \ref{subfig:ransac}; whereas Fig. \ref{subfig:gtm} depicts GTM result, as well as the result after a refinement of false positives from GTM result based on the area value and the direction of segments (cf. Fig. \ref{subfig:gtm_final}).

	\subsection{Transformation model estimation}
	The coordinates of the matched segment centers are then used to estimate the transformation model composed of parameters of the imaging camera pose. They are exterior orientation parameters, which are the position $(X_0, Y_0, Z_0)$ and orientation $(\omega, \phi, \kappa)$ of the camera when the image was acquired. In this paper, this estimation is carried out using Gold Standard algorithm detailed in \cite[p.187]{hartley2003multiple}.
	\begin{figure}[!t]
		\centering
		\subfloat[Overlapping of the back-projected LiDAR point cloud on the respective orthorectified optical image (data set no. 1), before (left) and after registration (right)]{\includegraphics[width=0.9\linewidth]{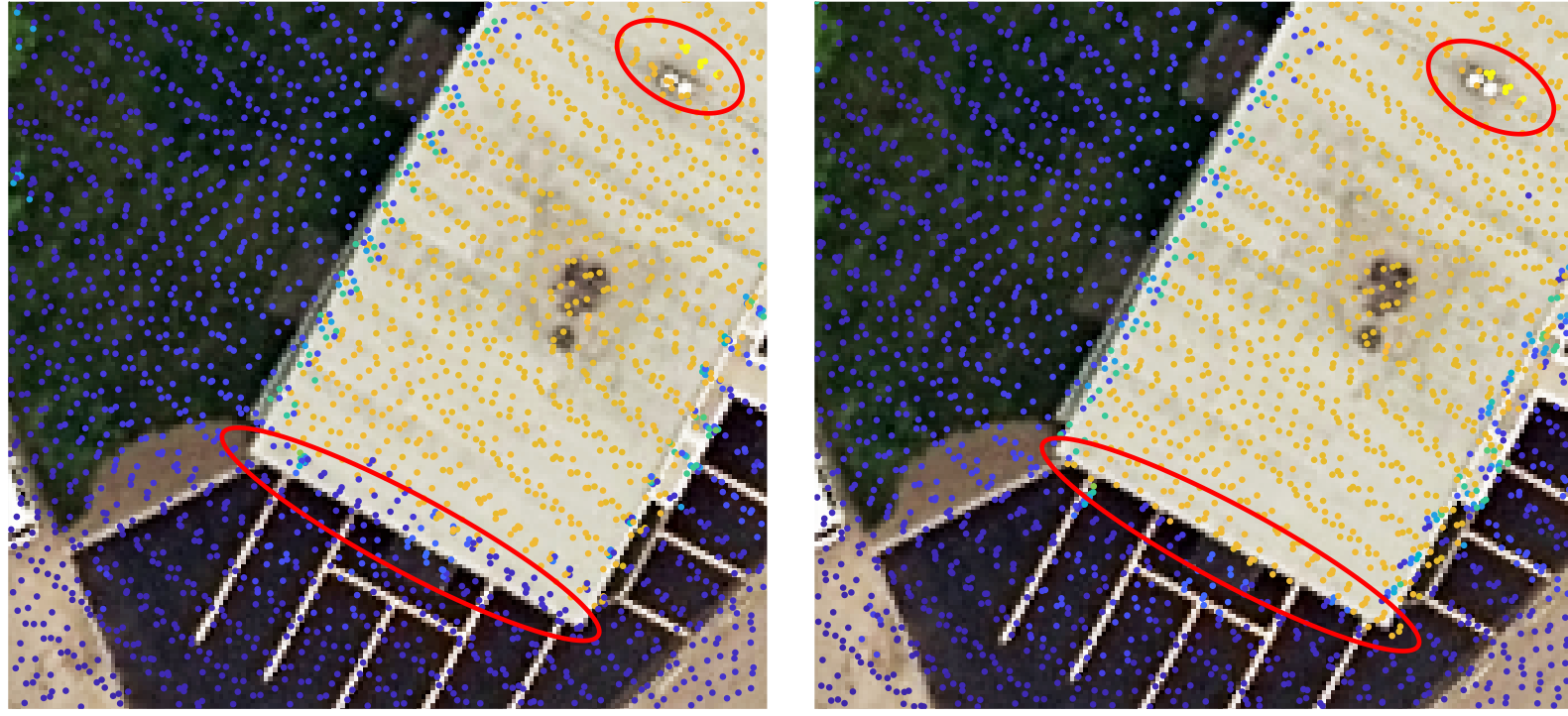}}
		\vspace{-0.25cm}
		\subfloat[Overlapping of the back-projected LiDAR point cloud on the respective satellite pansharpened image (data set no. 3), before (left) and after registration (right). Red arrows indicate displacements of some building corners for better understanding.]{\includegraphics[width=0.9\linewidth]{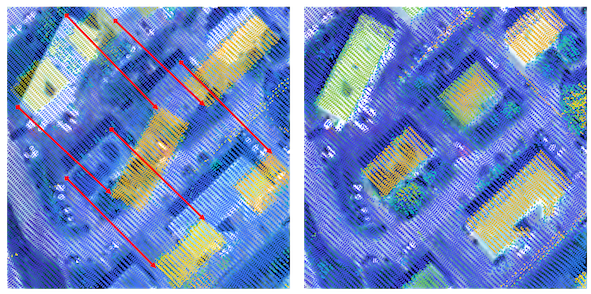}}
		\vspace{-0.25cm}
		\caption{\newtextt{Assessment of improved alignment} of the back-projected LiDAR point cloud overlapping on the respective image. }
		\label{fig:result}
	\end{figure}

	\section{Experimental results}
	\label{sec:results}
	\begin{table*}[!t]
		\centering
		\scalebox{0.8}{%
			\begin{tabular}{c|m{2.8cm}|p{2.9cm}|p{1.5cm}|p{3.5cm}|p{2.9cm}|p{1.5cm}}
				\hline 
				\multirow{2}{0.5cm}{\textbf{No.}} & \multirow{2}{*}{\textbf{Data type}} & \multirow{2}{*}{\textbf{Spectral resolution}} & \multirow{2}{1.4cm}{\textbf{Spatial resolution}}  & \multirow{2}{*}{\textbf{Acquisition time (season)}} & \multirow{2}{*}{\textbf{Geometry/Properties}} & \multirow{2}{1.8cm}{\textbf{Estimated relative shift}} \\
				&  & & &  & & \\
				\hline 
				\multirow{3}{*}{1} & \multirow{2}{*}{Aerial optical imagery} & \multirow{2}{2cm}{8 bits (RGBI)} & \multirow{2}{1.3cm}{15 cm} & \multirow{2}{*}{June 2016 (summer)} & $ \bullet $ Orthorectified & \multirow{3}{2cm}{1 - 2 m}\\
				&  & & &  & $ \bullet $ Georeferenced & \\[1pt]
				\cline{2-6}
				& LiDAR & 8 bits (Intensity) & 8 points/m$^2$ & May-Jun 2017 (summer) & Classified & \\
				
				\hline
				\multirow{3}{*}{2} & \multirow{2}{*}{Aerial optical imagery} & \multirow{2}{2cm}{8 bits (RGBI)} & \multirow{2}{1.3cm}{15 cm} & \multirow{2}{*}{Jul-Aug 2013 (summer)} & $ \bullet $ Central perspective & \multirow{3}{1.4cm}{2.5 - 10 m}\\
				& & & & & $ \bullet $ No georeferencing & \\
				\cline{2-6}
				& LiDAR & 8 bits (Intensity) & 2 points/m$^2$ & Oct-Nov 2011 (winter) & Classified & \\
				
				\hline 
				\multirow{3}{*}{3} & \multirow{2}{*}{Satellite imagery} & Panchromatic & 50 cm & \multirow{2}{*}{July 2015 (summer)} & $\bullet$ No orthorectification & \multirow{3}{*}{25 - 40 m} \\[0.5pt]
				\cline{3-4}
				& & \multirow{1}{*}{Multispectral (4 bands)} & 2 m & & $ \bullet $ Georeferenced & \\
				\cline{2-6}
				& LiDAR & 8 bits (Intensity) & 2 points/m$^2$ & Oct-Nov 2011 (winter) & Classified & \\
				\hline
		\end{tabular}}
		\caption{Details of datasets: LiDAR \textcopyright Ville de Qu\'{e}bec, aerial imagery \textcopyright Communaut\'{e} M\'{e}tropolitaine de Qu\'{e}bec, and satellite imagery \textcopyright Centre National d'\'{E}tudes Spatiales (France).}
		\label{tab:datasets}
	\end{table*}
	
	The proposed registration method has been tested using three different pairs of data sets, as described by Table \ref{tab:datasets}. Table \ref{tab:comparative} summarizes the results of building extraction and matching on selected areas, as the number of true positives (\newtextt{TP}, i.e. good extraction or matching), false alarms (\newtextt{FA}, i.e. wrongly extracted or matched), and misses (\newtextt{M}, i.e. buildings exist but not extracted or not matched).
	
	\begin{table}[!h]
		\centering
		\scalebox{0.8}{%
			\begin{tabular}{c|P{1.75cm}|P{1.75cm}||P{1.75cm}|P{1.75cm}}
				\hline 
				& Extracted from LiDAR data &  Extracted from image by mean shift & Matching result by RANSAC &  Matching result by GTM \\
				\hline
				\textbf{TP/FA/M} & 28/0/0 & 24/\textbf{21}/4 & 8/0/\textbf{12} & \textbf{19}/7/1 \\
				\hline
				\textbf{Precision} & 100\% & 53.33\% & 100\% & 73.08\% \\
				\hline
				\textbf{Recall} & \textbf{100\%} & \textbf{85.71\%} & 40\% & \textbf{95\%} \\
				\hline
		\end{tabular}}
		\vspace{-0.1cm}
		\caption{Performance of building extractions and matching algorithms on selected areas (28 buildings in total).} 
		\label{tab:comparative}
	\end{table}
	
	The overall results of the registration can be assessed through a reduction of the relative shift between data sets, measured from several manually selected control points, cf. Table \ref{tab:result}. This reduction is also demonstrated by  overlapped data sets before and after the registration on Fig. \ref{fig:result}. All full-scale color figures of this paper can be found on {\footnotesize\url{https://github.com/nthuy190991/igarss2019}}.
	\begin{table}[!h]
		\centering
		\scalebox{0.8}{%
			\begin{tabular}{c|P{1.5cm}|P{1.5cm}|P{1.5cm}}
				\hline 
				\multirow{2}{*}{\textbf{Data set}} & \multicolumn{3}{c}{\textbf{Average estimated relative shift}}\\ 
				\cline{2-4}
				& \textbf{Before} & \textbf{After} & \newtextt{\textbf{Gain}} \\
				\hline
				1 & 1.41 m & 0.49 m & 65.25\%\\
				\hline
				2 & 2.83 m & 1.32 m & 53.36\%\\
				\hline
				3 & 40.81 m & 1.75 m & 95.71\%\\
				\hline
		\end{tabular}}
		\vspace{-0.1cm}
		\caption{Average estimated relative shift between data sets, before and after the registration.}
		\label{tab:result}
	\end{table}
	
	\section{Conclusions and Perspectives}
	\label{sec:conclusions}
	In this paper, we present a dedicated registration approach between airborne LiDAR data and optical imagery which are not acquired from the same platform, \newtext{neither} with the same point of view \newtext{nor} the same spatial resolution. This approach focusing on extracting and matching building regions, allows reducing drastically the relative shifts between data sets, namely more than 50\% of the displacement between LiDAR data and aerial imagery and  95.71\% of the displacements between LiDAR data and satellite imagery. It also improves the alignment when overlapping the back-projected LiDAR point cloud on the optical image. Based on these results, a fine registration between these data sets could be applied, which is necessary to align them at an accuracy of 1-pixel level. This accuracy is required in order to fully benefit from the advantages of the two data sets and carry out a fusion providing a better completeness and a reduced uncertainty of the observed scene \cite{Parmehr2014}.

	\section{Acknowledgment}
The authors would like to thank the Centre G\'{e}oStat (Universit\'{e} Laval, QC, Canada), as well as Qu\'{e}bec City, Communaut\'{e} M\'{e}tropolitaine de Qu\'{e}bec (Canada), and Centre National d'Etudes Spatiales (France) for providing the data sets used in this work.
	
	
	\bibliographystyle{ieeetr}
	{\footnotesize
		\bibliography{reference.bib}
\end{document}